# Knowledge-graph based Proactive Dialogue Generation with Improved Meta-Learning

Hongcai Xu*, Junpeng Bao, Junqing Wang

Xi'an Jiaotong University, Xi'an, Shaanxi, 710049, P.R. China

Xajt1822@stu.xjtu.edu.cn

**Abstract.** Knowledge graph-based dialogue systems can narrow down knowledge candidates for generating informative and diverse responses with the use of prior information, e.g., triple attributes or graph paths. However, most current knowledge graph (KG) cover incomplete domain-specific knowledge. To overcome this drawback, we propose a knowledge graph based proactive dialogue generation model (KgDg) with three components, improved model-agnostic meta-learning algorithm (MAML), knowledge selection in knowledge triplets embedding, and knowledge aware proactive response generator. For knowledge triplets embedding and selection, we formulate it as a problem of sentence embedding to better capture semantic information. Our improved MAML algorithm is capable of learning general features from a limited number of knowledge graphs, which can also quickly adapt to dialogue generation with unseen knowledge triplets. Extensive experiments are conducted on a knowledge aware dialogue dataset (DuConv). The results show that KgDg adapts both fast and well to knowledge graph based dialogue generation and outperforms state-of-the-art baseline.

## 1 Introduction

Neural dialogue system is considered as a machine that can proactively converse with humans on a variety of topics [1, 2]. Recent years have seen some truly remarkable conversational AI systems on the market (e.g. Apple Siri, Amazon Alexa, Microsoft Cortana). However, these agents remain far from the stage where they can help users accomplish specific tasks such as making restaurant reservations and booking hotels and also keep humans engaged with natural and informative responses. The current research approaches, sequence to sequence models with attention of various kinds [3, 4], which directly map dialog history to the output responses, have shown promising results, but usually suffer from general responses (without useful information).

Recently, there are knowledge-aware conversation models [5, 6, 7] incorporating external knowledge in an end-to-end fashion to generate informational responses. The knowledge in these models may contain either unstructured texts [5] or domain-specific knowledge triplets [8, 9]. Nonetheless, these works are highly limited by the small-scale, domain-specific knowledge graphs. As a matter of fact, knowledge graphs are often incomplete and insufficient to cover full entities and relationships due to the human costs of collecting knowledge triplets. Meanwhile, they rarely analyze the coverage of KGs about different topics which contain the limited entities and relationships, and consider the embedding of KGs consistent to history dialogue utters. Therefore, how to generate knowledge aware response from limited resources is an essential task in KG-based dialog system. During the past few years, transfer learning [10], few-shot learning [11] and meta-learning [12, 13] have been introduced to solve such data scarcity problem. And current dialogue system usually respond to users' utters, rather than on its own initiatives. For example, as shown in Fig. 1, the knowledge graph corresponding to each Goal contains few entities and relationships, and for each turn in the conversation, the dialogue system proactively selects related knowledge triplets to construct responses.

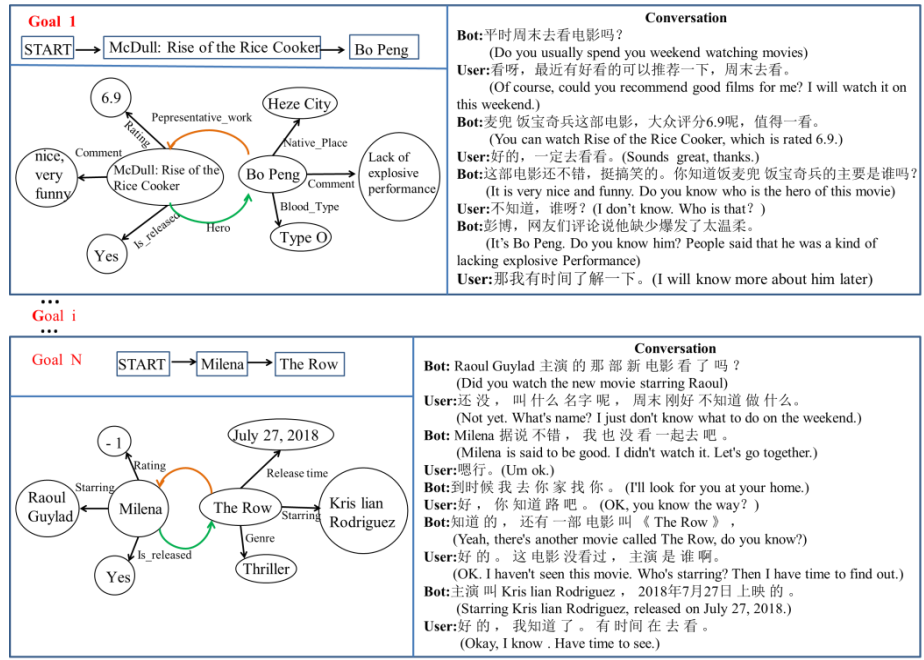

**Fig. 1.** Examples of KG based proactive conversation.

In this paper, we propose a conversational model which is an optimization-based meta-learning to perform KGs-based proactive dialogue generation and learn the way of KG embedding consistent to history dialogue. The goal of MAML is to find an initialization that can adjust to unseen knowledge triplets. By modifying the generation-based model (gen w/ klg) [14], a sequence to sequence (seq2seq) model with knowledge selector, we implement the MAML algorithm to achieve an optimal initialization and fine-tune the initialization towards unseen knowledge graphs with a minimal portion of dialog data.

To sum up, the contributions of this paper are as follows:

1. We propose an improved meta-learning algorithm for knowledge-aware dialogue generation. Since each knowledge-graph has a small number of entities and relationships, meta-learning can quickly accommodate to new knowledge dialogue tasks.

2. We propose an improved MAML algorithm which can fully utilize the information of support set and then update model weights to better learn the optimal initialization.

Furthermore, unlike existing models that use knowledge triplets (entities) separately and independently, our model treats each knowledge graph as a whole, therefore encoding more structured, connected semantic information in the graphs. Experimental results demonstrate that our proposed methods make full use of related knowledge to generate more diverse conversations.

## 2  Related Work& Background

### 2.1  Knowledge aware dialogue generation

In general, popular chit-chat systems only pay attention to dialogue history sentences, but do not consider the related knowledge in the model. Goal-oriented dialogue, such as restaurant booking [15], only utilizes the domain knowledge through API calls. Question answering systems [16, 17], considering structured knowledge triplets and unstructured texts, generate an answer based on a question rather than conversation history. Several recent knowledge-grounded conversation models incorporating unstructured texts [18, 20] and structured knowledge [9, 19, 21] have been presented in order to address the problem of generic response generation. Ghazvininejad et al. [18] utilized memory networks which store unstructured texts to generate engaging, meaningful or informative responses. Xu et al. [9] embedded structured domain-specific knowledge base into conversation generation model with a recall-gate mechanism. Niu et al. [19] incorporated a commonsense knowledge base into conversation generation with knowledge interpreter and knowledge aware generator. Tuan et al. [21] proposed an knowledge graph-based dialogue system that can zero-shot adapt to dynamic knowledge graphs. However, these works did not consider the limited number of knowledge triplets. By contrast, our model applies meta-learning algorithm to learn an initialization that quickly generates informative responses limited by new knowledge triplets and small-scale knowledge graph.

### 2.2  Meta-Learning

Meta-learning or learning-to-learn has recently received extensive interests and attention, which tries to adjust to new tasks quickly with few steps and little data. The key idea of meta-learning is to learn an optimal initialization. Recently, several successful applications of meta-learning have already been proposed, such as image classification [12], machine translation [23], neural conversation system [24, 25, 26], etc. There are three categories of meta-learning methods: (1) Metric-based meta-learning is about learning a metric space, which is used to compare testing tasks and training tasks [27, 28, 29, 30]. (2) Optimization-based meta-learning is about learning an optimal initialization that accommodates to new task quickly with limited training samples [22, 23, 24, 25, 26, 34]. (3) Policy-based meta-learning is about learning a meta-policy to update model parameters with relatively fewer training tasks [31, 32, 33]. In this paper, we propose a knowledge graph-based dialogue model based on MAML, which applies two-step gradient updates on DuConv to generate informative proactive dialogue. Because MAML is compatible with any neural conversation model optimized with gradient descent. More specifically, we focus on learning an optimal initialization that could adapt to new knowledge aware dialogue generation task.

## 3    Task Definition and Overview

For knowledge graph-based proactive dialogue generation model $M_\theta$, the goal of proactive dialogue is to naturally shift the topics twice, i.e., from "[start]" state to "topic a" then to "topic b". Two linked knowledge entities in knowledge graph refer to "topic a" and "topic b", which are used to construct the knowledge path [14], as shown in Fig.1. Given a history dialogue x and knowledge paths sampled from knowledge graph $\mathcal{G}$, the goal is to generate an informative response y consistent to knowledge entities and relationships. Our model based on MAML utilizes a set of tasks $\{\mathcal{T}_1,\ldots,\mathcal{T}_K\}$ to find the optimal model initialization adaptive to new task. The task $\mathcal{T}_i$ consists of a loss function $\mathcal{L}_i$ and a dataset $D_i$ that is further split into $D_i^{train}$ (support set) and $D_i^{test}$ (query set). $D_i^{train}$ is used for updating model parameters which will be utilized to calculate the loss $\mathcal{L}$ on $D_i^{test}$, and then compute the gradient based on $\mathcal{L}$ to update meta parameters. We have the training data $D_i^{train}$ consistent to each knowledge path, denoted as: $D_i^{train} = \{(x_n^k, y_n^k, \mathcal{G}^k), n = 1\ldots N\}, k = 1\ldots K$. We also denote the testing data $D_i^{test}$ different from training data: $D_i^{test} = \{(x_m^t, y_m^t, \mathcal{G}^t), m = 1\ldots M\}, t = 1\ldots T$. N and M are dialogue length. K and T refer to the number of knowledge path. The process of learning model parameters θ can be understood as [23]:

$$\theta_i^* = learn(\mathcal{T}_i; MetaLearn(\mathcal{T}_1,\ldots,\mathcal{T}_k)) \qquad (1)$$

where $\mathcal{T}_i$ is the target task. Hopefully, by exposing models to a variety of tasks, the models can learn new tasks in just a few steps and minimal amounts of data.

## 4    Meta-Learning for Knowledge graph-based Proactive dialogue generation

To the best of our knowledge, it is the first work which applies the meta-learning algorithm to knowledge graph-based proactive dialogue generation tasks. Our work analyses the limitations of small scale knowledge path as background knowledge, and then combines the meta-learning with the knowledge graph-based dialogue system to adjust to new knowledge path or knowledge triplets. In this section, we first describe the knowledge graph-based proactive dialogue model. Then, we explain how meta-learning combines dialogue model to generate better response and adapt to new knowledge triplets by leveraging only minimal amounts of conversational data.

### 4.1 Proactive Human-Machine Conversation Model

In knowledge graph-based proactive dialogue generation, we treat the proactive conversation model [14] as the base model $M_\theta$. The model consists of four modules: the history dialogue encoder, knowledge triplets encoder, the knowledge manager and the decoder. Each history dialogue context $X = \{x_1, x_2, ..., x_m\}$, along with the dialogue goal $G = \{g_1,...,g_n\}$, where $g_i$ is a knowledge path consisting of two topics (knowledge entities) and structuring as "[start]→ topic_a→ topic_b". For each turn in dialogue process, the proactive speaker selects knowledge triplets and generates response closer to the target topic. $\mathcal{G} = \{\mathcal{G}_1,..., \mathcal{G}_k\}$ refers to knowledge graph, where $\mathcal{G}_1$ is consistent to a goal $g_1$. $\mathcal{G}_1 = \{H, R, T\}$ is a single knowledge graph, in which H, T $\in \mathcal{V}$ (the set of entities or attributes) and R is a set of relationships. (h, r, t) is a knowledge triplets in knowledge graph. Given the X, $g_1$, and $\mathcal{G}_1$, these information is feed to an encoder which is bi-directional GRUs, and then a sequence $Y = \{y_1, y_2, ..., y_n\}$ is generated with a decoder, namely GRU with attention mechanism.

The knowledge graph-based dialogue model $M_\theta$ contains three loss functions (KLDivLoss, NLL Loss, and BOW Loss) [14]. They are defined as follows:

**KLDivLoss.** This loss aims to mimic human participation in knowledge selection. Two distributions, $p(k_i, x)$ and $p(k_i, x, y)$, are proposed, where $p(k_i, x)$ refers to knowledge reasoned by model, while $p(k_i, x, y)$ refers to knowledge reasoned by human. The formulas for these concepts are as follows:

$$p(k_i \mid x, y) = \frac{\exp(k_i \cdot MLP([x; y]))}{\sum_{j=1}^{N} \exp(k_j \cdot MLP([x; y]))} \quad (2)$$

$$p(k_i \mid x) = \frac{\exp(k_i \cdot x)}{\sum_{j=1}^{N} \exp(k_j \cdot x)} \quad (3)$$

$$\mathcal{L}_{KL}(\theta) = \sum_{i=1}^{N} p(k = k_i \mid x, y) \log \frac{p(k = k_i \mid x, y)}{p(k = k_i \mid x)} \quad (4)$$

**NLL Loss.** The Negative Log-Likelihood (NLL) $\mathcal{L}_{NLL}(\theta)$ is often used to measure the difference between the ground response and the response generated by our model, and evaluate how well our dialogue system generates dialogue.

$$\mathcal{L}_{NLL}(\theta) = -E_{k_i \sim p(k|x,y)} \sum_{t=1}^{m} \log p(y_t \mid y_{<t}, x, k_i) \quad (5)$$

**BOW Loss.** The BOW loss is intended to ensure the accuracy of the fused knowledge by enforcing the relevancy between the knowledge and the true response.

$$\mathcal{L}_{BOW}(\theta) = -E_{k_i \sim p(k|x,y)} \sum_{t=1}^{m} \log p(y_t \mid k_i) \quad (6)$$

In summary, the final loss of our generative model is:

$$\mathcal{L}(\theta) = \mathcal{L}_{KL}(\theta) + \mathcal{L}_{NLL}(\theta)\mathcal{L}_{BOW}(\theta) \tag{7}$$

### 4.2 Improved MAML for knowledge graph-based dialogue generation

In this section, we first describe the improved MAML algorithm and then introduce how to combine MAML with knowledge aware dialogue model, as illustrated in algorithm 1. An updating episode includes an inner gradient update process (Line5–Line 9) and a meta-update process (Line 11). The overview of the proposed meta-learning framework is shown in Fig. 2.

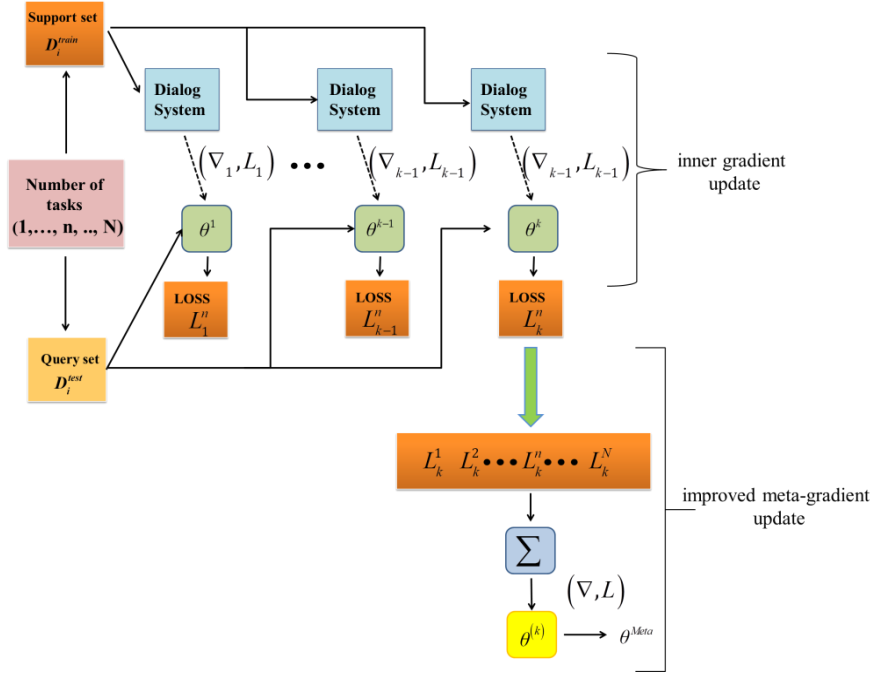

**Fig. 2.** An overview of our improved meta-learning framework.

We first introduce meta-learning stage. Section 4.1 describes the knowledge graph-based model $M_\theta$ with parameters θ. At each step, we first sample a batch of tasks containing support set $D_i^{train}$ and query set $D_i^{test}$, and then update the parameters by k(k≥1) gradient descent steps for each task $\mathcal{T}_i$ in the following equations.

$$\theta_i^k = \theta_i^{k-1} - \alpha \nabla_{\theta_i^{k-1}} \mathcal{L}_{\mathcal{T}_i}\left(M_{\theta_i^{k-1}}, D_1^{train}\right) \tag{8}$$

$$\mathcal{L}_{\mathcal{T}_i}(\theta) = \mathcal{L}_{KL}(\theta_i^k) + \mathcal{L}_{NLL}(\theta_i^k)\mathcal{L}_{BOW}(\theta_i^k) \tag{9}$$

where $\mathcal{L}_{\mathcal{T}_i}$ is the model loss function for task $\mathcal{T}_i$ defined in section 4.1, and α the learning rate. In this update procedure, the training data is from support set $D_i^{train}$.

The knowledge graph-based dialogue model parameters θ are then updated based

on the $\theta_i^k$:

$$\theta = MetaUpdate\left(\theta; \{\theta_i^k\}, D_i^{test}\right) \quad (10)$$

In the improved MetaUpdate stage, the knowledge aware dialogue data is from query set $D_i^{test}$.

MetaUpdate is the core step of learning-to-learn. The improved MAML is to learn the model parameters with the meta-objective function:

$$\min_{\theta^{(k)}} \sum_{T_i \sim p(T)} \mathcal{L}_{T_i}\left(M_{\theta_i^k}, D_i^{test}\right) = \min_{\theta^{(k)}} \sum_{T_i \sim p(T)} \mathcal{L}_{T_i}\left(M_{\theta_i^k = \theta_i^{k-1} - \alpha \nabla_{\theta_i^{k-1}} \mathcal{L}_{T_i}\left(M_{\theta_i^{k-1}}, D_i\right)}, D_i^{test}\right) \quad (11)$$

where $\theta^{(k)}$ refers to the model weights by k gradient descent steps for some batch of tasks.

The improved MetaUpdate of model $M_{\theta^{(k)}}$ is to update $\theta^{(k)}$ according to:

$$\theta^{(k)} \leftarrow \theta^{(k)} - \beta \sum_{T_i \sim p(T)} \nabla_{\theta^{(k)}} \mathcal{L}_{T_i}\left(M_{\theta_i^k}, D_i^{test}\right) \quad (12)$$

In order to fully utilize the knowledge of the learning process based on support set, we use the support set to update the parameters of the dialogue model in each num_task, and calculate the loss on the corresponding query set without updating the dialogue model.

We further introduce the real implementation of improved MAML in Fig. 3. For each task $\mathcal{T}_i$, in the inner gradient update, KgDg first drags θ via gradient descent to the direction (dotted arrow) reaching point $\theta^k$. Then in the improved meta-update (solid arrow), based on $\theta^k$, KgDg optimizes $\theta^{(k)}$ to reach the optimal point $\theta^{Meta}$.

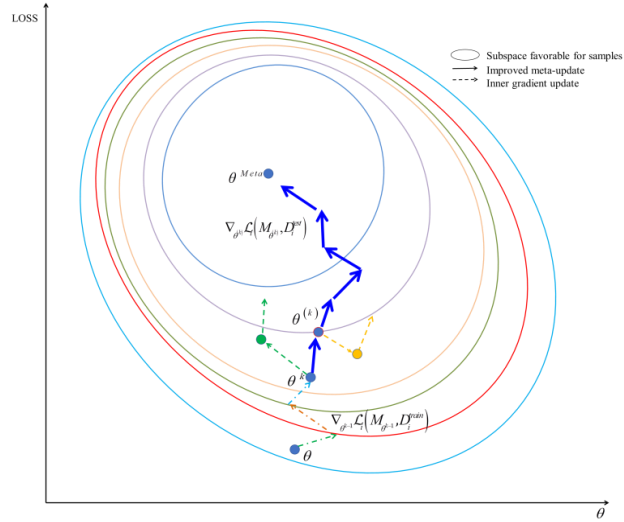

**Fig. 3.** Real implementation of improved MAML.

---

Algorithm 1 Training Algorithm of meta-learning model for knowledge graph-based dialogue generation ($M_\theta$, $\theta_0$, D, α, β)

---

    **Input:** $M_\theta, \theta_0, D$, α, β
    **Output:** $\theta^{Meta}$, optimal meta-learned model
1: Randomly initialize $\theta = \theta_0$
2: **while** not done **do**
3:   Sample batch of tasks $<\mathcal{T}_i>$ (num_tasks) from task set $\mathcal{T}$ ;
4:   **for all** $\mathcal{T}_i$ **do**
5:     Sample K samples $\{(x_n^1, y_n^1, \mathcal{G}^1),…,(x_n^K, y_n^K, \mathcal{G}^K)\}$ from $D_i^{train}$ for the inner gradient update;
6:     Compute updated model parameters with gradient descent. In inproved MAML, update the parameters by k(k≥1) gradient descent steps,
$$\theta^{(k)} := \theta^{(k-1)} - \alpha \nabla_{\theta^{(k-1)}} \mathcal{L}_i(M_{\theta}^{(k-1)}, D_i^{train});$$
7:     Sample k samples $\{(x_n^1, y_n^1, \mathcal{G}^1),…,(x_n^k, y_n^k, \mathcal{G}^k)\}$ from $D_i^{test}$ for the improved meta-gradient update;
8:   **end for**
9:   Improved Meta Update $\theta^{(k)} := \theta^{(k)} - \beta \nabla_{\theta^{(k)}} \sum_{\mathcal{T}_i \sim \mathcal{T}} \mathcal{L}_i(M_\theta^{(k)}, D_i^{test})$;
10: **end while**
11: Return $\theta^{Meta} = \theta^{(k)}$

---

## 5 Experiments

### 5.1 Experimental Setup

**Evaluation Metrics.** To verify whether KgDg can produce more consistent and informative responses with reference to the given history dialogue and knowledge triplets (even though not seen), we adopt several common metrics including BLEU1/2, PPL, F1,DISTINCT1/2 to automatically evaluate the fluency, relevance, diversity, etc. The BLEU score is used to determine whether the generated response is also part of the task. For each turn dialog, the entity F1 Score is used to check the completeness of tasks. PPL is a measurement of how well our model predicts a sample. DISTINCT score is utilized to measure the diversity of generated responses.

**Model and Learning.** We implemented Mθ by using a knowledge grounded neural seq2seq architecture (gen w/ klg) [14] with pre-trained word2vec embedding. We used the default setting of hyperparameters (embed_size =300, hidden_size = 300, max_vocab_size = 30000, inner update_step = 4, update_step_test = 10). We set nun_task to 5. k shot for support set is 8. k shot for query set is 14. batch size is 110 (5*(8+14)) . In meta-training, we used Adam for the inner loop and the outer loop with learning rate α = 0.0001 and β = 0.0001.

**Dataset.** DuConv dataset contains 270K utterances of 30K dialogues and about 3.6 million knowledge triplets. The dataset is divided into training set, validation set, and test set. The dialogue topics in DuConv are about movies and film stars. In training set and validation set, each sample includes Dialogue Goal, Background Knowledge and Conversation. In test set, each sample includes Dialogue Goal, Background Knowledge and History.

**Baselines.** To evaluate the effectiveness of the improved meta-learning algorithm, we compare KgDg with the following two baselines: a seq2seq model attending on knowledge graph is widely used in open-domain conversational systems. A knowledge-grounded model (gen w/ klg) [14] has the ability to proactively lead the conversation. To achieve better performances, the baseline normalized train/valid/test data by replacing the specific two topics in the knowledge path with "topic a" and "topic b" respectively. But, in actual conversation, human beings instinctively engage in conversation by understanding the semantics of topics or entities. So we did not normalize the dataset.

### 5.2 Results

As presented in Table 1, results show that KgDg significantly outperforms other baselines. For a fair comparison, we use models based on seq2seq architecture. We compare our model with its non-meta-learning counterpart utilizing PPL, F1, BLEU1, BLEU2, DISTINCT 1, and DISTINCT 2. Our model achieves 1.1% and 3% gains on the BLEU1 and BLEU2 score, respectively. Our approach performs slightly better on PPL and F1 compared to gen w/o klg and gen w/ klg, yet worse than S2S w/o klg, where gen w/o klg refers to the proactive conversation model, S2S stands for the sequence-to-sequence model, and klg stands for knowledge. The experimental results reveal that our dialogue system outperforms the strong gen w/o klg baseline across different settings, indicating the better effectiveness of the improved meta-learning algorithm for knowledge graph-based proactive dialogue generation tasks.

**Table 1.** Evaluation results.

| Methods    | PPL   | F1    | BLEU1 | BLEU2 | DISTINCT 1 | DISTINCT 2 |
|------------|-------|-------|-------|-------|------------|------------|
| S2S w/o klg | 20.16 | 26.43 | 0.187 | 0.100 | 0.032      | 0.088      |
| S2S w/ klg  | **13.53** | 32.19 | 0.226 | 0.140 | **0.064**  | **0.168**  |
| Gen w/o klg | 20.3  | 28.52 | 0.29  | 0.154 | 0.032      | 0.075      |
| Gen w/ klg  | 27.3  | 36.21 | 0.32  | 0.169 | 0.049      | 0.144      |
| KgDg       | 20.1  | **36.72** | **0.33** | **0.200** | 0.051  | 0.147      |

As presented in Table 2, with the proposed improved meta-learning and fine-tuning, the final knowledge graph based dialogue system trained using only a fraction of all available training examples is particularly strong in performance. As the adaptation size decreases from 76800 to 300, 900, 3000, the performance of Gen w/ klg drops quickly, while KgDg performs stably well.

**Table 2.** Results for adapting to knowledge graph based dialogue generation with adaptation size 300, 900, 3000.

| Methods | Adaptation size | PPL | F1 | BLEU1 | BLEU2 | DISTINCT 1 | DISTINCT 2 |
|---|---|---|---|---|---|---|---|
| Gen w/klg | 300 | 114.48 | 19.72 | 0.201 | 0.092 | 0.029 | 0.042 |
| KgDg | 300 | **90.99** | **23.85** | **0.244** | **0.183** | **0.060** | 0.031 |
| Gen w/klg | 900 | 68.90 | 28.23 | 0.249 | 0.130 | 0.010 | 0.080 |
| KgDg | 900 | **41.45** | **29.56** | **0.288** | **0.149** | **0.024** | 0.050 |
| Gen w/klg | 3000 | 49.91 | 31.83 | 0.312 | 0.167 | 0.032 | 0.077 |
| KgDg | 3000 | **22.49** | **34.17** | **0.313** | **0.188** | **0.049** | **0.124** |

**Table 3.** Validation results during first six epochs.

| Epoch | Step | PPL KgDg | PPL Gen w/klg | LOSS KgDg | LOSS Gen w/klg | Accurate rate KgDg | Accurate rate Gen w/klg |
|---|---|---|---|---|---|---|---|
| 1 | 200 | 42.16 | 30128.476 | 130.00 | 212.72 | 0.36 | 0 |
| | 400 | 33.15 | 30134.013 | 125.68 | 211.78 | 0.39 | 0 |
| | 600 | 28.79 | 30128.579 | 123.10 | 210.72 | 0.41 | 0 |
| 2 | 200 | 26.18 | 30129.142 | 121.53 | 209.58 | 0.42 | 0 |
| | 400 | 24.13 | 30125.228 | 120.10 | 209.12 | 0.44 | 0 |
| | 600 | 22.95 | 30130.29 | 118.95 | 208.75 | 0.45 | 0 |
| 3 | 200 | 22.00 | 30136.63 | 118.25 | 208.26 | 0.45 | 0 |
| | 400 | 21.03 | 30136.02 | 117.48 | 208.00 | 0.46 | 0 |
| | 600 | 20.68 | 30136.16 | 116.91 | 207.76 | 0.46 | 0 |
| 4 | 200 | 19.86 | 30137.13 | 116.28 | 207.46 | 0.46 | 0 |
| | 400 | 19.32 | 30137.41 | 115.72 | 207.26 | 0.47 | 0 |
| | 600 | 18.92 | 30139.58 | 11.53 | 207.09 | 0.47 | 0 |
| 5 | 200 | 18.47 | 30137.77 | 115.08 | 206.89 | 0.48 | 0 |
| | 400 | 18.37 | 30137.52 | 114.71 | 206.78 | 0.47 | 0 |
| | 600 | 18.13 | 30139.94 | 114.64 | 206.65 | 0.48 | 0 |
| 6 | 200 | 18.00 | 128.626 | 114.45 | 132.67 | 0.48 | 0.21 |
| | 400 | 17.79 | 69.647 | 113.98 | 124.32 | 0.48 | 0.32 |
| | 600 | 17.83 | 50.2 | 114.12 | 119.85 | 0.48 | 0.36 |

To assess the quickly adaptation of the proposed method, we present the validation results in Table 3. We can see that the proposed approach not only obtain better adaption quickly but also achieves better diversity, consistency and compliance. These results demonstrate that the improved meta-learning approach is more robust to the drop in the size of the knowledge graph based dialogue generation task's training set.

To further investigate the adaptation process, we present the performance curves of Genw/ klg and KgDg in Fig.4. We observe that the two curves of PPL scores fall quickly, and then increase. This result shows that our model is able to quickly adapt to generate informative responses with a few training epochs compared to Genw/ klg model. Similar to training other neural dialogue systems, meta-learning still requires early-stopping to avoid overfitting.

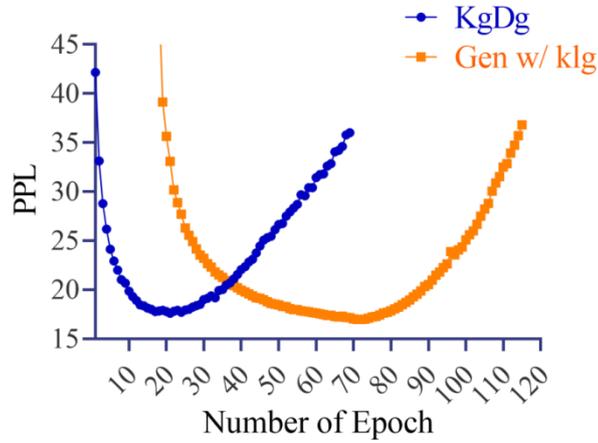

**Fig. 4.** The learning curves of PPL scores on the validation task.

We evaluate the performance of our improved MAML and the Genw/ klg model on BLEU1 and BLEU2 scores on the validation task which indicate the consistence about the generated sentences. From Fig.5, we can see that our model converges much faster especially during the first 5 epochs, while the Genw/ klg model that is trained using standard supervised learning is unable to adequately adapt to generate consistent responses with a few epochs. We also find the BLEU1 and BLEU2 scores of Genw/ klg model almost equal to zero during the first 5 epochs. These results demonstrate that the improved MAML trained model is better than the baseline.

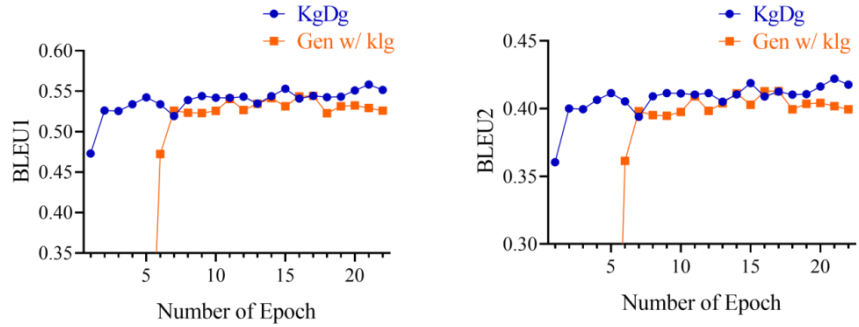

**Fig. 5.** Comparing BLEU scores between KgDg and Genw/ klg.

Furthermore, in the Fig.6, we present both models' results on DISTINCT1 and DISTINCT2 scores which count the number of distinct unigrams and bigrams, respectively. The KgDg curves continue to rise steeply, while the Genw/ klg curves increase slowly and are always lower than KgDg curves. These results further demonstrate improved MAML algorithm is more suitable for knowledge graph-based dialogue generation tasks. In summary, our models perform better than the baseline, indicating our improved meta-learning algorithms can find a balance between general knowledge aware dialogue generation information and adaptive to task-specific information.

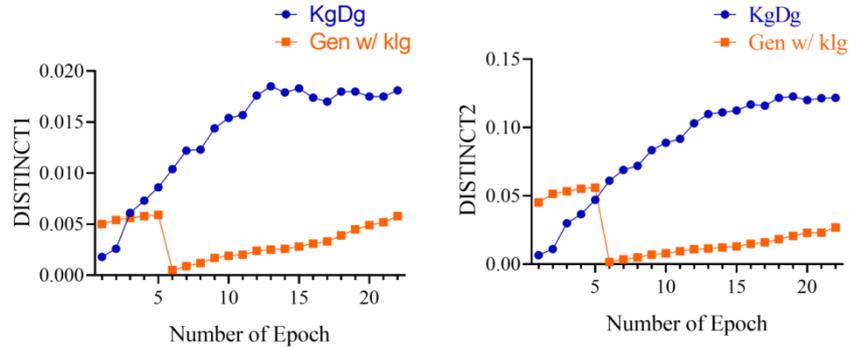

**Fig. 6.** Comparing DISTINCT scores between KgDg and Genw/ klg.

### 5.3 Case Study

Fig.7 shows the generated sentences by our models, given the conversation goal and the related knowledge. We can observe that our knowledge graph based dialogue generator with improved meta-leaning can choose appropriate knowledge triplets for diverse conversation generation.

| | |
|---|---|
| Goal | START → Milena → The Row |
| Knowledge | ["Milena", "发布日期信息", "20年前上映"], ("Milena", "Release Date Information", "Released 20 years ago") ["Milena", "口碑", "口碑很差"],("Milena", "Word of mouth", "Poor reputation") ["Milena", "导演", "维拉·贝蒙"],("Milena", "Director", "Villa Baymon") ["Milena", "主演", "Nick Mancuso"], ("Milena", "Starring", "Nick Mancuso ")["The Row", "发布日期信息", "上个月上映"],(The Row ", "Release Date Information", "Released Last Month" ) |
| Knowledge-graph based Proactive Dialogue model (KgDg) | **Bot:**我看了一部电影，是20年前上映。<br>(I saw a movie that was released 20 years ago.)<br>**User:**是什么？(What is it?)<br>**Bot:**电影叫 Milena，口碑较差 。<br>(The movie is called Milena, and bad word of mouth.)<br>**User:**是吗？还可以吧！(Really? Okay!)<br>**Bot:**这部电影的导演是维拉·贝蒙,主演是Nick Mancuso,可以看一下。<br>(The director of this movie is Vera Belmont, Starring Nick Mancuso ,Take a look.)<br>**User:**是个很有才华的人，我很喜欢。<br>(A very talented person, I like the man very much.)<br>**Bot:**还有一部电影叫 The Row ,上个月前就已经上映了,你可以看看。<br>(There is also a movie called The Row, which was released last month. You can watch it.) |

**Fig. 7.** Example of the response generation.

## 6   Conclusion and future work

In this paper, we propose a knowledge-aware dialog generation method based on improved meta-learning, called KgDg, which uses improved meta-gradients to find the optimal initialization weights that are more robust to different knowledge paths. We evaluate our model on DuConv dataset. According to the results presented above, KgDg reaches the state-of-the-art performance compared with the baseline. In future works, we plan to apply meta-learning to lifelong interactive knowledge learning systems.

**Acknowledgment**. This work was supported by 2019 basic scientific research business fee (free exploration and innovation-students) project(1191329120).